\documentclass[%
 aip,
 amsmath,amssymb,
 reprint,%
]{revtex4-1}

\usepackage{graphicx}
\usepackage{dcolumn}
\usepackage{bm}

\usepackage[utf8]{inputenc}
\usepackage[T1]{fontenc}
\usepackage{mathptmx}
\usepackage{etoolbox}
\usepackage[english]{babel}

\usepackage[usenames,dvipsnames]{color}

\makeatletter
\def\@email#1#2{%
 \endgroup
 \patchcmd{\titleblock@produce}
  {\frontmatter@RRAPformat}
  {\frontmatter@RRAPformat{\produce@RRAP{*#1\href{mailto:#2}{#2}}}\frontmatter@RRAPformat}
  {}{}
}%
\makeatother
\begin{document}

\preprint{AIP/123-QED}

\title[Transfer Learning for Molecular Property Predictions from Small Data Sets]{Transfer Learning for Molecular Property Predictions from Small Data Sets}
\author{Thorren Kirschbaum}
 \affiliation
{Theory of Electron Dynamics and Spectroscopy, Helmholtz-Zentrum Berlin für Materialien und Energie GmbH, Hahn-Meitner-Platz 1, 14109 Berlin, Germany}
\author{Annika Bande}
 \email{annika.bande@helmholtz-berlin.de}
\affiliation
{Theory of Electron Dynamics and Spectroscopy, Helmholtz-Zentrum Berlin für Materialien und Energie GmbH, Hahn-Meitner-Platz 1, 14109 Berlin, Germany}
\affiliation{Institute of Inorganic Chemistry, Leibniz University Hannover, Callinstr. 9, 30167 Hannover, Germany}

\date{\today}

\begin{abstract}
Machine learning has emerged as a new tool in chemistry to bypass expensive experiments or quantum-chemical calculations, for example, in high-throughput screening applications. However, many machine learning studies rely on small data sets, making it difficult to efficiently implement powerful deep learning architectures such as message passing neural networks. In this study, we benchmark common machine learning models for the prediction of molecular properties on two small data sets, for which the best results are obtained with the message passing neural network PaiNN, as well as SOAP molecular descriptors concatenated to a set of simple molecular descriptors tailored to gradient boosting with regression trees. 
To further improve the predictive capabilities of PaiNN, we present a transfer learning strategy that uses large data sets to pre-train the respective models and allows to obtain more accurate models after fine-tuning on the original data sets. The pre-training labels are obtained from computationally cheap ab initio or semi-empirical models and 

both data sets are normalized to mean zero and standard deviation one to align the labels' distributions. 
This study covers two small chemistry data sets, the Harvard Organic Photovoltaics data set (HOPV, HOMO-LUMO-gaps), for which excellent results are obtained, and on the Freesolv data set (solvation energies), where this method is less successful, probably due to a complex underlying learning task and the dissimilar methods used to obtain pre-training and fine-tuning labels. Finally, we find that for the HOPV data set, the final training results do not improve monotonically with the size of the pre-training data set, but pre-training with fewer data points can lead to more biased pre-trained models and higher accuracy after fine-tuning.
\end{abstract}

\maketitle

\section{Introduction}


In the realm of molecular sciences, predicting the properties of molecules stands as a pivotal pursuit across numerous domains, including pharmaceuticals, materials science, and environmental studies. Traditionally, this endeavor has been tackled by quantum chemistry, relying on sophisticated theoretical frameworks and computational simulations to unravel the intricate behaviors of molecules. However, the advent of machine learning (ML) has ushered in a paradigm shift, offering an alternative approach that leverages data-driven methodologies efficiently compute molecular properties.\cite{montavon_machine_2013, hansen_machine_2015, dral_quantum_2020, fung_benchmarking_2021, singh_graph_2022}

One notable application of ML in this domain is its integration into high throughput screening workflows, where it replaces conventional quantum chemistry methods.\cite{dral_quantum_2020} Through the judicious application of ML algorithms, researchers can expedite the process of identifying promising molecules for drug discovery or materials development, thereby accelerating the pace of scientific innovation. However, the powerful deep neural networks (NNs) often used in such approaches usually need large amounts of training data to finally make accurate predictions.

Machine learning algorithms, particularly deep NNs, have exhibited remarkable success in discerning intricate patterns and relationships within data sets of substantial size. In recent developments, the message passing neural network PaiNN has demonstrated highest performance in predicting molecular properties, achieving state-of-the-art results across various datasets.\cite{schutt_equivariant_2021, kirschbaum_machine_2023} Another prevalent approach in ML for molecular sciences entails the utilization of tailored molecular descriptors, such as SOAP,\cite{bartok_representing_2013, parsaeifard_assessment_2021} in conjunction with traditional ML models like kernel ridge regression (KRR).
In general, as a rule of thumb, deep learning architectures are to be preferred when $>$ 10$^3$–10$^4$ training points are available, and different approaches, such as KRR, should be considered otherwise.\cite{unke_machine_2021} Accordingly, molecular fingerprinting methods are commonly used in junction with KRR for regression on small data sets.\cite{zhao_effect_2020, zhao_performance_2022, kirschbaum_machine_2023, odral_ai_2024} Yet, in the context of molecular property prediction, data sets often fall short regarding the typically required large amounts of data for deep NNs. This mismatch often presents a formidable obstacle, impeding the efficacy of ML in scenarios characterized by small data sets, which are common in practical applications.\cite{ramakrishnan_big_2015, zhang_strategy_2018, mazouin_selected_2022}
Bridging this divide necessitates innovative strategies and methodologies tailored to the unique challenges posed by small data sets. In this paper, we present a transfer learning strategy to address this pressing need.

In transfer learning, the knowledge learned in one setting is used to improve performance on a related target task.\cite{zhuang2020comprehensive} In a first step, known as pre-training, an NN is trained on a large set of data that is closely related to the target task, yielding an NN with pre-optimized parameters. In the second step, termed fine-tuning, the same model is re-trained using data of the target task. Now, since the model is already close to the optimal set of parameters after the pre-training, much less data is needed during fine-tuning to obtain an accurate model for the target task. This strategy has proven useful in various fields, including image classification\cite{saenko2010computer} and captioning,\cite{vinyals2015show} gaming strategies,\cite{sharmacl} and social network analysis.\cite{tang2016transfer} 
In the physical sciences, a range of studies recently investigated the use of transfer learning to improve atomic force predictions for molecular dynamics simulations.\cite{smith_approaching_2019, kaser_machine_2020, zheng2021artificial, kaser2022transfer, gao2022supervised, zhang2022dpa, chen2023data, zaverkin_transfer_2023}
Moreover, some previous studies have used transfer learning to improve property predictions of molecules and materials.\cite{li2022improving, shui2022injecting} 
For example, Su \textit{et al.} recently presented a transfer learning strategy for predicting frontier orbital energies of organic materials based on DFT data obtained using two different functionals for generating pre-training and fine-tuning labels, respectively.\cite{su_deep_2023} Furthermore, several studies have explored the use of transfer learning for regression on both HOPV\cite{zhao_effect_2020, eibeck_predicting_2021, munshi_transfer_2021, li_improving_2022, moore_deep_2022} and Freesolv.\cite{vermeire_transfer_2021, li_improving_2022, zhang_accurate_2022}
For such tasks, Sun \textit{et al.} found that supervised pre-training generally helps, if the pre-training labels are closely aligned with the downstream tasks, but the success of efficient transfer learning strongly depends on the employed data sets, ML models, and their hyperparameters.\cite{sun2022does} 
Hoffmann \textit{et al.} further showed that increasing both the size of the pre-training and the fine-tuning data sets linearly improved the prediction results in the transfer learning settings investigated by them.\cite{hoffmann_transfer_2023} However, as pointed out by Dral, relatively little research is done in exploring the advantages and limitations of transfer learning in the context of repurposing the models for different levels and molecules.\cite{odral_ai_2024}

In this study, we start by undertaking a rigorous benchmarking, evaluating standard ML techniques in the context of small molecular data sets. In particular, we compare the use of SOAP descriptors tailored to a standard ML algorithm (KRR, NN, or gradient boosting) to the message passing neural network PaiNN for regression. The comparison is done both on the Harvard Organic Photovoltaics (HOPV) data set,\cite{lopez_harvard_2016} containing photovoltaic properties of 350 organic molecules, and on the Freesolv data set,\cite{duarte_ramos_matos_approaches_2017} which contains solvation energies of 643 small organic molecules.

Secondly, we propose a transfer learning strategy as a means of augmenting predictive modeling with small data sets. 
It involves leveraging computations from computationally cheap ab initio or semi-empirical models to generate pre-training labels for PaiNN. 


\section{Methods}\label{sec:Methods}

In this study, we use two small common physical chemistry data sets, the Harvard Organic Photovoltaics (HOPV) data set\cite{lopez_harvard_2016} and the Freesolv data set.\cite{duarte_ramos_matos_approaches_2017} The HOPV data set contains 350 molecules and their photovoltaic data obtained from both experiments and quantum-chemical calculations. In this study, we focus on the HOMO-LUMO gaps obtained from PBE0/def2-SVP density functional theory, which is a common \textit{ab initio} method used for standard chemistry data sets.\cite{adamo_toward_1999, perdew_rationale_1996, schafer_fully_1992, weigend_balanced_2005, rupp_fast_2012, stuke_atomic_2020, kirschbaum_machine_2023}
Note that the version of the HOPV data set used by us (HOPV\_15\_revised\_2) contains a single molecule with an Se atom, which we regarded as a structural outlier and therefore removed from the data set prior to training.
The Freesolv data set contains experimental and calculated solvation energies of 643 small molecules,\cite{duarte_ramos_matos_approaches_2017} from which we use the values obtained from "alchemical" calculations based on molecular dynamics simulations by the Mobley group.\cite{mobley_freesolv_2014} 

For benchmarking standard ML methods with HOPV and Freesolv, we employed the PaiNN implementation of the \textit{schnetpack} Python library.\cite{schutt_schnetpack_2019}
PaiNN was recently proposed as an extension of the popular MPNN SchNet.\cite{schutt_equivariant_2021} In the PaiNN architecture, molecular atoms are represented as graph nodes, and interactions between atoms are represented as undirected edges between the nodes. Each atom (node) is described by atomic scalar features $s$, and each interaction (edge) as edge vector features $h$ that are updated using stacked message-passing and update blocks. After the message phasing phase, the final predictions are obtained after summing or averaging over the final node descriptors.\cite{schutt_equivariant_2021}

We further used the SOAP implementation of the \textit{dscribe} library\cite{dscribe} in junction with either \textit{sklearn}'s GradientBoostingRegressor (GBoost), kernel ridge regression, or a simple feed-forward NN. The GBoost algorithm profited from an additional set of simple molecular descriptors concatenated to the SOAP descriptor. For NN learning, to limit the size of the NNs, the sparse SOAP descriptors were compressed by principal component analysis, retaining 99.999 \% of the variance.\cite{parsaeifard_assessment_2021, kirschbaum_machine_2023}
The hyperparameters of PaiNN were optimized using an iterative grid search, and the hyperparameters of the SOAP-ML models were optimized using Bayesian optimization. For PaiNN, a learning rate decay and early stopping were employed, and results were averaged over five runs. In all cases, 20 \% of the data was used for testing, and for the NNs, 20 \% of the data was used for validation and early stopping. The final model hyperparameters are listed in the appendix.

In the second part of this study, we present a method to improve the predictions of the message passing neural network PaiNN,\cite{schutt_equivariant_2021} which was also identified as the best ML model for the data sets at hand (see Results and Discussion), via transfer learning.
For this, we need (i) a (small) molecular data set with high-accuracy data of the target property ($y_{\textrm{true}}$), (ii) a large data set of molecular structures to be used for pre-training, and (iii) a computationally inexpensive method from the field of quantum chemistry to compute the target property for the pre-training data. Cheap ab initio methods, like density functional theory in the linear density approximation (LDA-DFT), or semi-empirical methods typically yield results which are qualitatively correct, but quantitatively far off.\cite{thiel_semiempirical_2014} 
To correct for the biases inherent in the results obtained from these methods, we normalize both data sets to have mean 0 and standard deviation 1 before NN training.

\begin{figure}[!htb]
\includegraphics[width = 0.45\textwidth]{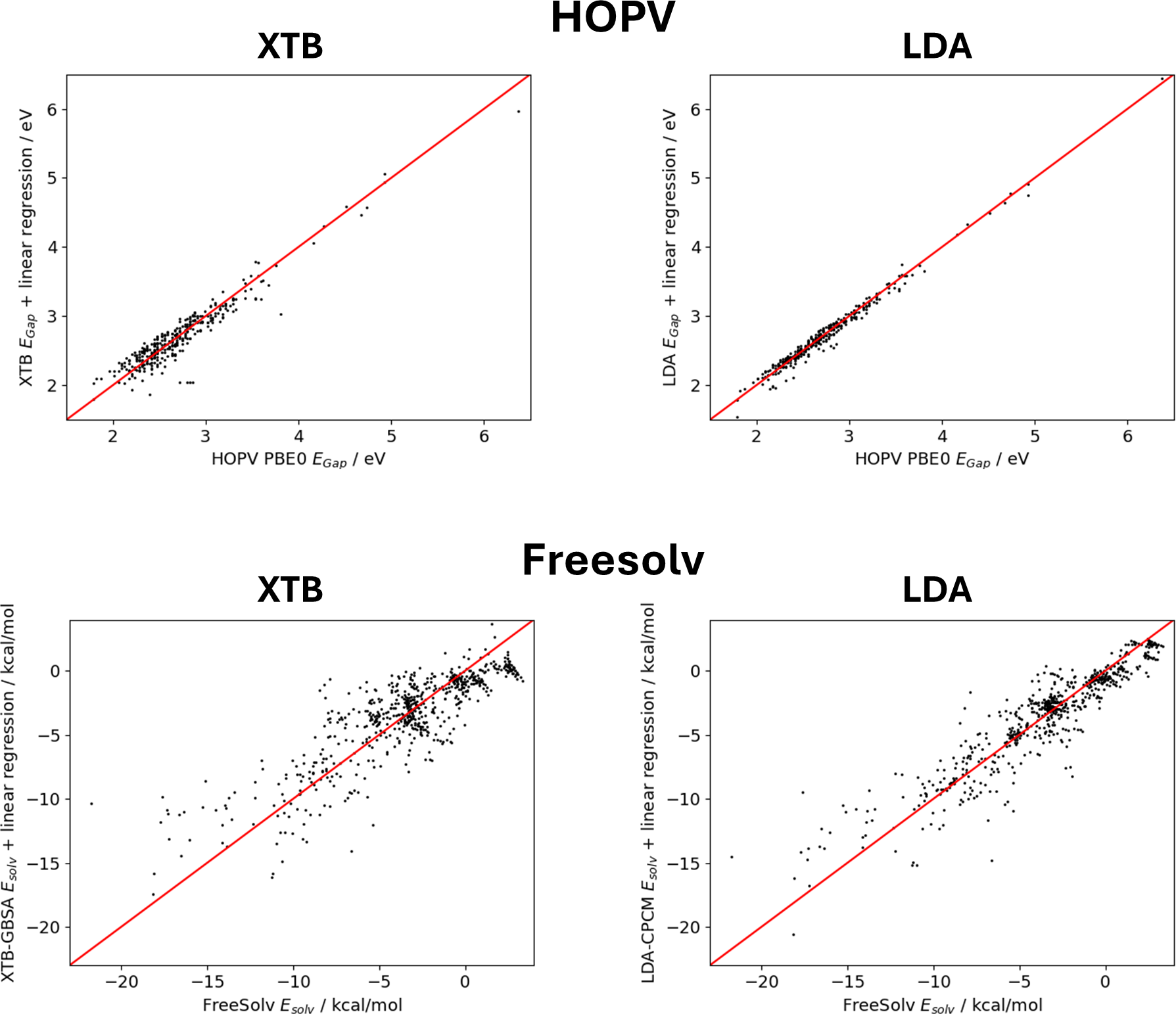}
\caption{Scatter plots of the original data plotted against data obtained from XTB (left) or LDA (right) plus linear regression for the data sets HOPV (HOMO-LUMO-gaps, top) and Freesolv (solvation energies, bottom).}
\label{fig:fits}
\end{figure}

In this study, we use the molecular structures of the OE62 data set as pre-training structures for the HOPV data set, and the molecular structures of the QM9 data set as pre-training structures for Freesolv (ii).
The OE62 data set contains crystal structures of 62 k medium-sized organic molecules,\cite{stuke_atomic_2020} and the QM9 data set contains DFT-optimized structures of 134 k small molecules.\cite{ramakrishnan_quantum_2014} To obtain pre-training labels for these data sets (iii), HOMO-LUMO gaps and solvation energies were calculated with LDA-DFT using the ORCA software,\cite{neese_orca_2020} and tight-binding density functional theory (DFTB) using the GFN2-xTB method\cite{bannwarth_gfn2-xtbaccurate_2019} implemented in the XTB software.\cite{bannwarth_extended_2021} Solvation energies were obtained as the difference in single point energy of the molecule in vacuum and the molecule with an implicit solvent model for water (conductor-like polarizable continuum model, CPCM in ORCA,\cite{barone_quantum_1998} and generalized Born model with surface area contributions, GBSA in XTB,\cite{onufriev_exploring_2004, sigalov_analytical_2006, lange_improving_2012} respectively). 
Note that we do not use the properties originally published with the respective data sets QM9 and OE62 during pre-training.
To illustrate the data quality obtained from LDA and XTB computations, we display the scatter plots of the original HOPV and Freesolv data, respectively, against the data obtained from LDA and XTB, respectively, corrected by linear regression (Fig. \ref{fig:fits}). These plots clearly show that the obtained HOMO-LUMO gaps are very close to the original HOPV data, but the solvation energies are quite dissimilar to the original Freesolv data. Furthermore, for both data sets, the properties obtained from LDA are closer to the original data those obtained from XTB.
The mean absolute errors of the fits are:
0.05 eV (HOPV LDA), 0.12 eV (HOPV XTB), 1.1 kcal/mol (Freesolv LDA), 1.7 kcal/mol (Freesolv XTB).

On average, HOMO-LUMO gap computations took approximately 106 s (LDA-DFT), and 0.22 s (XTB) per data point in OE62, respectively.
Solvation energy computations took approximately 70 s (LDA-DFT), and 0.15 s (XTB) per data point in QM9, respectively, and 1.33 s (XTB) per data point in OE62.

Pre-training on QM9 data was done for 200 epochs (100 k training, 10 k validation, 20 k test data points). 
Pre-training on OE62 data was done for 250 epochs for LDA and XTB data (50 k training, 5 k validation, 6 k test data points).
For each method, three models were trained with different random seeds and the best model as evaluated on the test subset was selected for fine-tuning.

\section{Results and Discussion}

\subsection{ML benchmarks}

We start by benchmarking standard chemistry ML models for regression on the two small, but high-quality data sets under investigation, HOPV\cite{lopez_harvard_2016} and Freesolv,\cite{duarte_ramos_matos_approaches_2017} listed under (i) in the Methods section. The results are summarized in Table \ref{tab:ML_benchmark} as mean absolute errors (MAEs, root mean squared errors (RMSE) are given in parenthesis). For both data sets, very low errors are obtained with PaiNN, yielding MAEs of 0.01 eV for HOPV and 0.56 kcal/mol for Freesolv, respectively. For HOPV, the SOAP+SD+GBoost model yields an MAE of 0.19 eV, while 0.86 kcal/mol are obtained for regression on Freesolv. The results obtained with these algorithms are qualitatively similar to those obtained in previous ML studies on HOPV\cite{paul_property_2019, zhao_effect_2020, munshi_transfer_2021, eibeck_predicting_2021, fare_multi-fidelity_2022, moore_deep_2022, li_improving_2022} and Freesolv.\cite{hutchinson_solvent-specific_2019, scheen_hybrid_2020, weinreich_machine_2021, borgis_accurate_2021, vermeire_transfer_2021, gao_graphical_2021, pathak_learning_2021, chung_group_2022, li_improving_2022, low_explainable_2022, zhang_accurate_2022, zhang_machine_2023}
The remaining models SOAP+NN and SOAP+KRR perform considerably worse on both data sets, in direct comparison SOAP+NN yielding better results for HOPV while SOAP+KRR yields better results for Freesolv. 


\begin{table}[!htb]
\caption{Mean absolute errors for HOPV HOMO-LUMO gaps (in eV) and Freesolv solvation energies (in kcal/mol) from four different models. The first three models use SOAP molecular descriptors connected to either KRR, a dense NN, or GBoost. The GBoost model uses an additional set of simple molecular descriptors (SD) concatenated with the SOAP descriptors. The final model is the message passing NN PaiNN.
Uncertainties are measured in RMSE (in parentheses). Best MAE results are bold.}

\begin{tabular}{l|r|r}
\textbf{Model} & \textbf{\begin{tabular}[c]{@{}l@{}}HOPV \\ MAE (RMSE) / eV\end{tabular}} & \textbf{\begin{tabular}[c]{@{}l@{}}Freesolv\\ MAE (RMSE) / kcal/mol\end{tabular}} \\
\hline
SOAP+KRR & 0.31 (0.46) & 1.7 (2.2) \\
SOAP+NN & 0.25 (0.33) & 2.1 (3.2) \\
SOAP+SD+GBoost & 0.19 (0.30) & 0.86 (1.30) \\
PaiNN & \textbf{0.01 (0.01)} & \textbf{0.56 (0.86)} \\
\end{tabular}
\label{tab:ML_benchmark}
\end{table}

\subsection{Transfer Learning}

\begin{figure}[!htb]
\includegraphics[width = 0.445\textwidth]{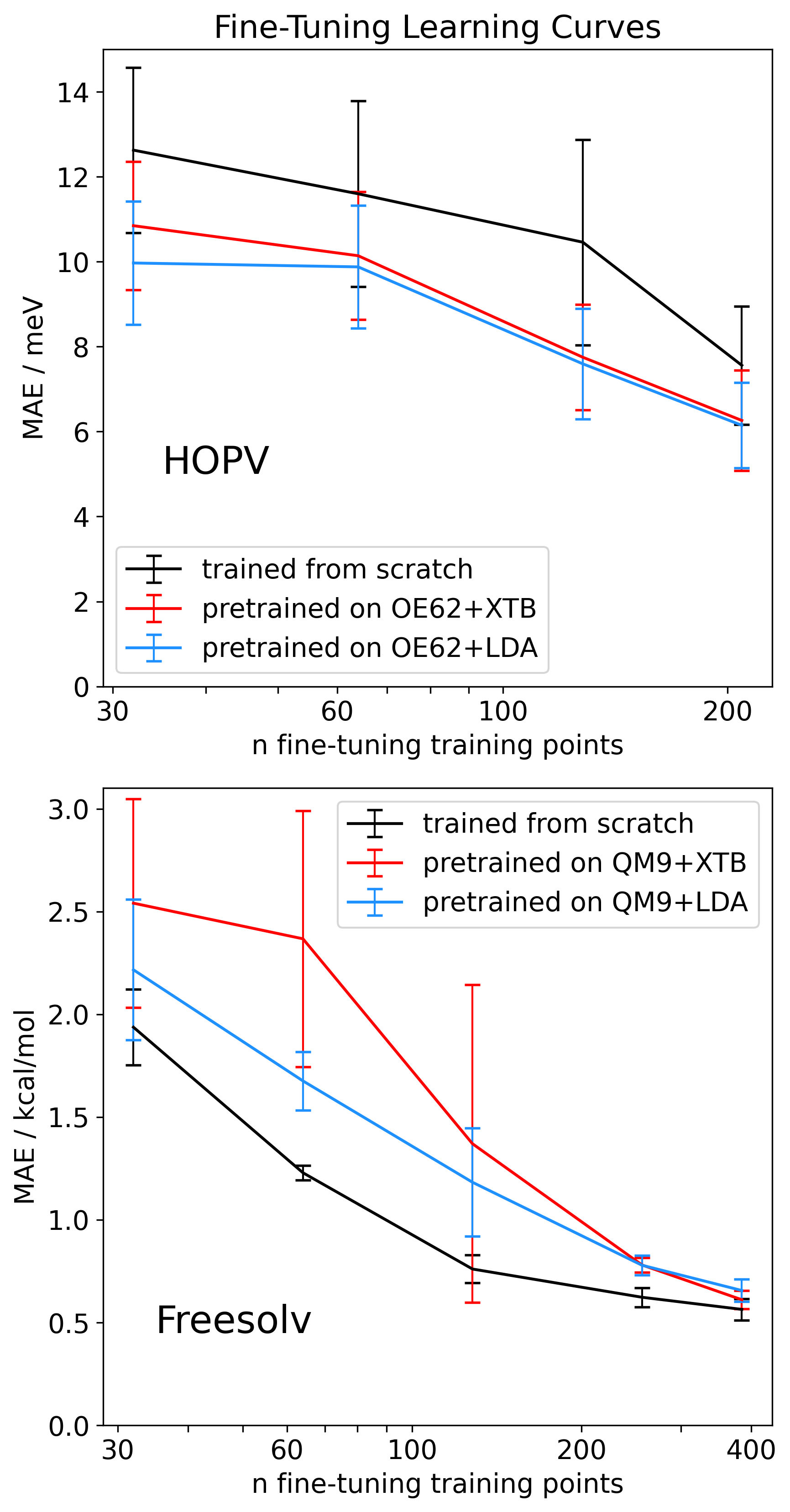}
\caption{Fine-tuning learning curves for PaiNN training on HOPV (top) and Freesolv (bottom), after training from scratch (black) or after pre-training on OE62 or QM9 data, respectively, with labels obtained from XTB (red) and LDA-DFT (blue). The MAE (mean and standard deviation over five runs) is plotted against the number of training examples used for fine-tuning (log scale).}
\label{fig:lc1}
\end{figure}

\begin{figure}[!htb]
\includegraphics[width = 0.45\textwidth]{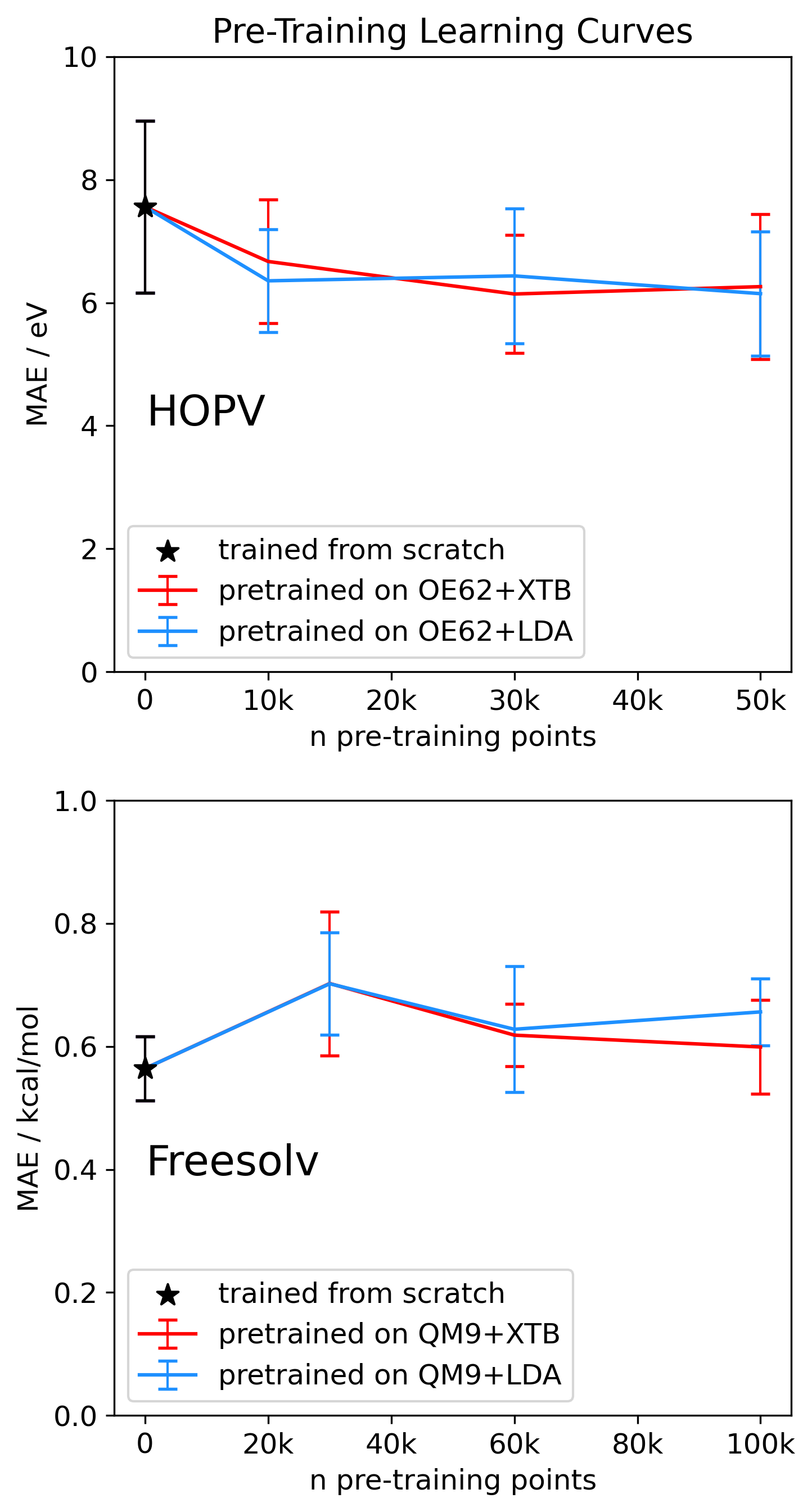}
\caption{Pre-training learning curves for PaiNN training on HOPV (top) and Freesolv (bottom), after training from scratch (black star) or after pre-training on OE62 or QM9 data, respectively, with labels obtained XTB (red) and LDA-DFT (blue). The MAE (mean and standard deviation over five runs) is plotted against the number of training examples used for pre-training.}
\label{fig:lc2}
\end{figure}

We use cheap models from quantum chemistry (LDA-DFT and XTB-DFTB) normalized to the same mean (zero) and standard deviation (one) as the fine-tuning data to obtain labels for pre-training the PaiNN models (details in the Methods section). For HOPV, consisting of medium-sized organic molecules, we use the structures of the OE62 data set for pre-training, which contains ca. 62 k diverse structures of medium-sized organic molecules. For Freesolv, which on the other hand consists of small organic molecules, we use the structures of the QM9 data set for pre-training, which contains ca. 132 k organic molecules with up to nine heavy (non-H) atoms. Tables \ref{tab:ML_HOPV} and \ref{tab:ML_Freesolv} summarize the results obtained for regression on the HOPV and Freesolv data set, respectively, after PaiNN training from scratch (cf. Table \ref{tab:ML_benchmark}), or after pre-training on data labeled with either XTB or LDA, respectively. The learning curves displaying the MAE vs. the number of fine-tuning training points are shown in Figure \ref{fig:lc1} (here, pre-training is done on the full respective pre-training data set), and the learning curves displaying the MAE vs. the number of pre-training points are shown in Figure \ref{fig:lc2} (here, fine-tuning is done on the full respective fine-tuning data set), each for the HOPV set on top and the Freesolv set in the bottom. All reported MAEs are obtained as the mean over five independent runs, and the error bars indicate the MAEs' standard deviations.

\begin{table}[!htb]
\caption{MAEs for HOPV HOMO-LUMO gaps (in eV) from PaiNN obtained after training from scratch, or after pre-training on OE62 data with labels obtained from XTB and LDA, respectively. 
Uncertainties are measured in RMSE (in parentheses). Best MAE results are bold.}

\begin{tabular}{l|r}
\textbf{Pre-training} & \textbf{\begin{tabular}[c]{@{}l@{}}HOPV \\ MAE (RMSE) / meV\end{tabular}}  \\
\hline
None & 7.6 (11.9)  \\
OE62+XTB & 6.3 (9.7) \\
OE62+LDA & \textbf{6.1 (9.6)} \\
\end{tabular}
\label{tab:ML_HOPV}
\end{table}

\begin{table}[!htb]
\caption{MAEs for Freesolv solvation energies (in kcal/mol) from PaiNN obtained after training from scratch, or after pre-training on OE62 data with labels obtained from XTB and LDA, respectively. 
Uncertainties are measured in RMSE (in parentheses). Best MAE results are bold.}

\begin{tabular}{l|r}
\textbf{Pre-training} & \textbf{\begin{tabular}[c]{@{}l@{}}Freesolv \\ MAE (RMSE) / kcal/mol\end{tabular}}  \\
\hline
None & \textbf{0.56 (0.86)}  \\
QM9+XTB & 0.61 (1.05) \\
QM9+LDA & 0.66 (1.10) \\
\end{tabular}
\label{tab:ML_Freesolv}
\end{table}

For HOPV (Table \ref{tab:ML_HOPV}), pre-training results in an improved learning performance. The MAE is reduced from 7.6 meV (training from scratch) to 6.3 meV after pre-training on OE62+XTB, and further to 6.1 meV eV after pre-training on OE62+LDA.
The fine-tuning learning curves for HOPV (Figure \ref{fig:lc1}, top) show that this improvement of ca. 1.5 meV over training from scratch (black curve) after pre-training on OE62+XTB (red) or OE62+LDA (blue) is relatively stable, independent of the number of HOPV fine-tuning training points. Also, the standard deviation of the MAEs is reduced after pre-training on either data set.

The HOPV pre-training learning curves (Figure \ref{fig:lc2}, top) confirm that for all pre-training data set sizes tested, the labels obtained from the QC models lead to a higher accuracy as compared to training from scratch. Interestingly, increasing the size of the pre-training data set does not monotonically improve the final training results. Instead, the final training errors after pre-training on 10k data points do not strongly improve when more pre-training data (up to 50k data points) is used.
This indicates that a pre-trained model that is too strongly fitted on the pre-training task may be worse for transfer learning than a more biased model that only learned to grasp high-level concepts from a limited amount of training data, which contradicts the findings in the recent study of Hoffmann \textit{et al.}\cite{hoffmann_transfer_2023}

Regression learning on Freesolv, on the other hand, is not improved by pre-training the PaiNN models (Table \ref{tab:ML_Freesolv}). Instead, the final training MAEs are increased from 0.56 kcal/mol (training from scratch) to 0.61 kcal/mol (XTB), and 0.66 kcal/mol (LDA). The fine-tuning learning curves (Figure \ref{fig:lc1}, bottom) show that no improvement is achieved throughout, independent of the size of the pre-training data set. Also, there is no clear trend in how the size of the pre-training data set influences the result of the subsequent fine-tuning. In addition, the MAE standard deviations increase when pre-training on either data set.

As discussed before, the regression task on Freesolv might be more complex as compared to regression on HOPV. The methods used for producing the respective pre-training labels (QC methods) are quite similar to the method used for generating HOPV data (standard DFT),\cite{lopez_harvard_2016} but quite dissimilar to the method used for generating Freesolv data ("alchemical" calculations based on molecular dynamics simulations, see also the scatter plots in Figure \ref{fig:fits}).\cite{mobley_freesolv_2014} These factors may be the reason for why the employed transfer learning strategy does work very well for use with the HOPV data set, but not for Freesolv.

Another hypothesis for why transfer learning might fail in certain scenarios is that the underlying regression task is too easy, and therefore no additional pre-training is needed.\cite{sun2022does} However, previous results for regression on molecular energies on the QM9 data set (free energy $G$, internal energy $U$, enthalpy $H$) using PaiNN indicated that MAEs as low as ca. 0.14 kcal/mol can be obtained on a sufficiently large data set, which is considerably lower than the MAEs obtained here.\cite{schutt_equivariant_2021} Also, these quantities ($G$, $U$, $H$) can easily be calculated from only the respective molecular structure, but solvation energy computations additionally need to take into account the shape and intermolecular interactions of the respective solvation shell, thus adding another level of complexity. Therefore, we anticipate that the regression task on Freesolv is not too easy for transfer learning.

In an attempt to obtain pre-trained models that are more robust and therefore allegedly more suitable for transfer learning, we added additional data points to the respective pre-training data sets. For this, we used XTB to generate additional labels for QM9 (HOMO-LUMO gaps) and OE62 (solvation energies). However, we found that pre-training on a joint OE62 and QM9 data set did not improve the final fine-tuning results as compared to pre-training on OE62 or QM9 data only, respectively, for both HOPV and Freesolv training. This is surprising, because NNs usually get more accurate when increasing the amount of training data, and new dissimilar data points can help to create more robust models. Here, instead, the introduction of more data points that are dissimilar to the data contained in the final fine-tuning data set apparently impedes the learning process. In particular, for Freesolv, the underlying problem of having a complex learning task and dissimilar methods for generating pre-training and fine-tuning labels could not be solved by augmenting the pre-training data set. This observation is in line with previous work on transfer learning in chemistry, which stated that the pre-training and fine-tuning data sets should be as closely aligned as possible to obtain the best results.\cite{li2022improving}

Following this idea, we aimed to instead make the pre-training and fine-tuning data sets more similar to each other: The HOPV data set consists almost entirely of S-functionalized aromatic molecules, while the OE62 data set used for pre-training contains a much more diverse set of molecules, including more structural variety and a wider range of atomic species. Accordingly, we filtered all molecules from the OE62 pre-training data set containing any of the elements As, Se, Br, Te or I, and furthermore removed all molecules that do not contain any sulfur atoms, thereby reducing the data set size from 61,489 to 41,487 molecules. Pre-training with 20 k less molecules from the filtered OE62 data set resulted in a slight loss of performance compared to pre-training of the full OE62+XTB data set (+0.01 eV MAE), which is approximately the same result as obtained after non-selective filtering of the pre-training data set (+0.02 eV MAE, cf. Figure \ref{fig:lc2}). This indicates that increased similarity between pre-training and fine-tuning structures might slightly improve the learning performance by yielding more tailored pre-trained models, but the effect found here is very small.

Finally, previous works have used discriminative fine-tuning to further optimize transfer learning.\cite{zaverkin_transfer_2023} In this method, the learning rate during fine-tuning is adapted layer-wise, such that the first layers have a much smaller learning rate than the final layers of the model. The idea is that the first layers, which learn to grasp fundamental features of the molecular data, will be sufficiently trained during pre-training, while the last layers, which derive the final predictions, will need more fine-tuning. We implemented discriminative fine-tuning for PaiNN, following the approach outlined by Zaverkin \textit{et al.}\cite{zaverkin_transfer_2023} Here, the learning rate (lr) for layer n$-$1 was adjusted as lr$_{n-1}$ = lr$_{n}$/5 with lr = 5e-4 being the learning rate of the last layer. However, we found that this method did not yield any improvements over non-discriminative fine-tuning for transfer learning with PaiNN. In contrast, we found that in general, modifications of the learning rate and learning rate decay had a significant impact on the learning performance, and sticking to the parameters identified by the initial hyperparameter sweep continued to yield the best results.

\section{Conclusions}

This study investigated the use of common ML algorithms for use with small data sets, focusing on HOPV (350 data points) and Freesolv (643 data points). We found that for these data sets, both the MPNN PaiNN, and gradient boosting with regression trees operating on SOAP molecular descriptors concatenated to a set of simple molecular descriptors yield accurate results, clearly outperforming the SOAP+NN and SOAP+KRR architectures.

Furthermore, we presented a method for data-efficient ML on small physical chemistry data sets using transfer learning which we tested on the two data sets HOPV and Freesolv. The pre-training data was obtained from computationally cheap QC calculations corrected by simple linear regression calibrated on the fine-tuning data set. This method utilizes the fact that many cheap ab initio or semi-empirical techniques yield quantitatively incorrect but qualitatively correct data, which can easily be corrected via linear regression. For the HOPV data set, the introduced method outperforms transfer learning with labels obtained from an ML model. Here, pre-training with labels computed by either XTB or LDA result in very similar learning performances, but XTB computations were much faster than LDA computations. Accordingly, labelling the pre-training data with XTB should be preferred. 

However, the same method did not yield improved learning results for regression on the Freesolv data set. We suppose that this may be due to the fact that the underlying learning task is much more complex and the data used for pre-training is too dissimilar to the data contained in Freesolv. 

Furthermore, the ML results obtained for HOPV indicate that pre-training with a limited amount of data points may be beneficial, since the obtained models are slightly more biased and less strongly fitted to the pre-training task. Furthermore, we found that augmenting the pre-training data sets with additional structures that are less similar to the fine-tuning data does not improve the learning performance. Finally, for the models trained here, sticking to the hyperparameters identified by the initial hyperparameter sweeps was more efficient than adapting the learning rates as done in discriminative fine-tuning. However, given the limited scope of this study, further research will be needed to investigate whether these findings can be generalized to other data sets as well.

In conclusion, the results obtained from this exploratory study suggest that for chemistry ML tasks that use small data sets, an effective transfer learning strategy should use pre-training on molecules that are similar to the molecules of the downstream task, a labelling method that is closely aligned to the method used for labelling the fine-tuning data, and an optimized training set size for the pre-training. Future studies may investigate whether these findings apply to other data sets as well.

\begin{acknowledgments}
TK and AB acknowledge support from the Helmholtz Einstein International Berlin Research School in Data Science (HEIBRiDS). 
Computing resources were kindly provided by the Freie Universität Berlin hpc cluster Curta\cite{bennett_curta_2020} and by the Helmholtz-Zentrum Dresden-Rossendorf. The introduction was in part formulated with the help of ChatGPT 3.5.
\end{acknowledgments}

\section*{Data Availability Statement}
Data openly available in a public repository:
The data underlying this study are openly available at https://github.com/ThorrenKirschbaum/TransferLearningPaiNN.

\appendix*
\section{Machine Learning Model Hyperparameters}

We here provide details of the ML models used for regression on both HOPV and Freesolv (\textit{cf.} Table \ref{tab:ML_benchmark}).

The final SOAP+KRR models used the following hyperparameters:\\
HOPV: $r_{cut}^{SOAP} = 5.63$, $n_{max}^{SOAP} = 7$, $l_{max}^{SOAP} = 7$, $alpha_{KRR} = 1.12$, kernel: cosine.\\
Freesolv: $r_{cut}^{SOAP} = 9.53$, $n_{max}^{SOAP} = 8$, $l_{max}^{SOAP} = 8$, $alpha_{KRR} = 0.59$, kernel: = cosine.

In the SOAP+NN model, the neural network consists of densely connected layers with steadily decreasing numbers of neurons, including batch normalization and a dropout layer with $p=0.3$ after the first layer. Here, 20 \% of the training data was used for validation, and early stopping was employed. The final models used the following hyperparameters:\\
HOPV: $r_{cut}^{SOAP} = 7.22$, $n_{max}^{SOAP} = 8$, $l_{max}^{SOAP} = 8$, $n_{layers} = 2$, $lr = 0.00860$, nonlinearity: swish.\\
Freesolv: $r_{cut}^{SOAP} = 9.94$, $n_{max}^{SOAP} = 7$, $l_{max}^{SOAP} = 7$, $n_{layers} = 4$, $lr = 0.00627$, nonlin.: swish.

In the SOAP+SD+GBoost model, the following simple molecular descriptors were employed: total number of atoms, number of atoms for each atom type in the data set, one-hot-encoding for each atom type (present/not present in the respective molecule), mean and standard deviation of the carbon-carbon-distance of neighboring carbon atoms within the molecule.
Using these descriptors with the KRR or NN did not yield any improvement for the respective learning algorithm.
The final SOAP+SD+GBoost models used the following hyperparameters:\\
HOPV: $r_{cut}^{SOAP} = 9.71$, $n_{max}^{SOAP} = 5$, $l_{max}^{SOAP} = 8$, $lr_{GBoost} = 0.059$,  $n\_estimators_{GBoost} = 295$,  $max\_depth_{GBoost} = 4$.\\
Freesolv: $r_{cut}^{SOAP} = 6.85$, $n_{max}^{SOAP} = 4$, $l_{max}^{SOAP} = 4$, $lr_{GBoost} = 0.250$,  $n\_estimators_{GBoost} = 158$,  $max\_depth_{GBoost} = 4$.

The final PaiNN models used the following hyperparameters:\\
HOPV: $radial\_basis = GaussianRBF$, $r_{cut} = 5.0$, $n\_rbf = 20$, $n\_atom\_basis = 30$,  $n\_interactions = 3$, $cutoff\_fn = CosineCutoff$, $lr = 5e-4$, $lr\_decay\_factor =  0.5$, $lr\_decay\_patience = 5$, $batchsize = 10$.\\
Freesolv: $radial\_basis = GaussianRBF$, $r_{cut} = 8.0$, $n\_rbf = 30$, $n\_atom\_basis = 200$,  $n\_interactions = 3$, $cutoff\_fn = CosineCutoff$, $lr = 5e-4$, $lr\_decay\_factor =  0.5$, $lr\_decay\_patience = 5$, $batchsize = 10$.\\

\bibliography{references}

\end{document}